\documentclass[11pt]{article}
\usepackage{acl2016}
\usepackage{times}
\usepackage{url}
\usepackage{latexsym}
\usepackage{amsmath}
\usepackage{amsfonts}
\usepackage{graphicx}
\usepackage{booktabs}
\usepackage{pbox}
\usepackage{multirow}

\DeclareMathOperator*{\argmax}{argmax} 

\aclfinalcopy 

\title{A Joint Model for Word Embedding and Word Morphology}

\author{Kris Cao and Marek Rei\\
  Computer Lab \\
  University of Cambridge \\
  United Kingdom \\
  {\tt kc391@cam.ac.uk} \\
}

\date{08/06/2016}

\begin{document}
\maketitle
\begin{abstract}
  This paper presents a joint model for performing unsupervised morphological analysis on words, and learning a character-level composition function from morphemes to word embeddings. Our model splits individual words into segments, and weights each segment according to its ability to predict context words. Our morphological analysis is comparable to dedicated morphological analyzers at the task of morpheme boundary recovery, and also performs better than word-based embedding models at the task of syntactic analogy answering. Finally, we show that incorporating morphology explicitly into character-level models help them produce embeddings for unseen words which correlate better with human judgments.
\end{abstract}

\section{Introduction}
Word embedding models associate each word in a corpus with a vector in a semantic space. These vectors can either be learnt to optimize performance in a downstream task \cite{Bengio,Collobert} or learnt via the distributional hypothesis: words with similar contexts have similar meanings  \cite{harris,word2vec1}. Current word embedding models treat words as atomic. However, words follow a power law distribution \cite{zipf}, and word embedding models suffer from the problem of sparsity: a word like `unbelievableness' does not appear at all in the first 17 million words of Wikipedia, even though it is derived from common morphemes. This leads to three problems: 

\begin{enumerate}
    \item word representations decline in quality for rarely observed words \cite{bullinaria}.
    \item word embedding models handle out-of-vocabulary words badly, typically as a single `OOV' token.
    \item the word distribution has a long tail, and many parameters are needed to capture all of the words in a corpus (for an embedding size of 300 with a vocabulary of 10k words, 3 million parameters are needed)
\end{enumerate}

One approach to smooth word distributions is to operate on the smallest meaningful semantic unit, the morpheme \cite{lazaridou,Botha2014}. However, previous work on the morpheme level has all used external morphological analyzers. These require a separate pre-processing step, and cannot be adapted to suit the problem at hand.

Another is to operate on the smallest orthographic unit, the character \cite{wangling,yoonkim}. However, the link between shape and meaning is often complicated \cite{saussure}, as alphabetic characters carry no inherent semantic meaning. To account for this, the model has to learn complicated dependencies between strings of characters to accurately capture word meaning. We hypothesize that explicitly introducing morphology into character-level models can help them learn morphological features, and hence word meaning.

In this paper, we introduce a word embedding model that jointly learns word morphology and word embeddings. To the best of our knowledge, this is the first word embedding model that learns morphology as part of the model. Our guiding intuition is that the words with the same stem have similar contexts. Thus, when considering word segments in terms of context-predictive power, the segment corresponding to the stem will have the most weight.

Our model `reads' the word and outputs a sequence of word segments. We weight each segment, and then combine the segments to obtain the final word representation. These representations are trained to predict context words, as this has been shown to give word representations which capture word semantics well \cite{word2vec2}. As the root morpheme has the most context-predictive power, we expect our model to assign high weight to this segment, thereby learning to separate root+affix structures.

One exciting feature of character-level models is their ability to represent open-vocabulary words. After training, they can predict a vector for any word, not just words that they have seen before. Our model has an advantage in that it can split unknown words into known and unknown components. Hence, it can potentially generalise better over seen morphemes and words and apply existing knowledge to new cases.

To evaluate our model, we evaluate its use as a morphological analyzer (\S\ref{morphology}), test how well it learns word semantics, including for unseen words (\S\ref{wordsim}), and examine the structure of the embedding space (\S\ref{syntactic}).

\section{Related Work}
While words are often treated as the fundamental unit of language, they are in fact themselves compositional. The smallest unit of semantics is the morpheme, while the smallest unit of orthography is the grapheme, or character. Both have been used as a method to go beyond word-level models.

\subsection{Morphemic analysis and semantics}
As word semantics is compositional, one might ask whether it is possible to learn morpheme representations, and compose them to obtain good word representations. Lazaridou et al. \shortcite{lazaridou} demonstrated precisely this: one can derive good representations of morphemes distributionally, and apply tools from compositional distributional semantics to obtain good word representations. Luong et al. \shortcite{luong} also trained a morphological composition model based on recursive neural networks. Botha and Blunsom \shortcite{Botha2014} built a language model incorporating morphemes, and demonstrated improvements in language modelling and in machine translation. All of these approaches incorporated external morphological knowledge, either in the form of gold standard morphological analyses such as CELEX \cite{celex} or an external morphological analyzer such as Morfessor \cite{morfessor}.

Unsupervised morphology induction aims to decide whether two words are morphologically related or to generate a morphological analysis for a word \cite{goldwater,Goldsmith}. While they may use semantic insights to perform the morphological analysis \cite{soricut}, they typically are not concerned with obtaining a semantic representation for morphemes, nor of the resulting word.

\subsection{Character-level models}
Another approach to go beyond words is based on on character-level neural network models. Both recurrent and convolutional architectures for deriving word representations from characters have been used, and results in downstream tasks such as language modelling and POS tagging have been promising, with reductions in word perplexity for language modelling and state-of-the-art English POS tagging accuracy \cite{wangling,yoonkim}. Ballesteros et al. \shortcite{ballesteros} train a character-level model for parsing. Zhang et al. \shortcite{zhang} do away with words completely, and train a convolutional neural network to do text classification directly from characters.

Excitingly, character-level models seem to capture morphological effects. Examining nearest neighbours of morphologically complex words in character-aware models often shows other words with the same morphology \cite{wangling,yoonkim}. Furthermore, morphosyntactic features such as capitalization and suffix information have long been used in tasks such as POS tagging \cite{wenduan,toutanova}. By explicitly modelling these features, one might expect good performance gains in many NLP tasks.

What is less clear is how well these models learn word semantics. Classical word embedding models seem to capture word semantics, and the nearest neighbours of a given word are typically semantically related words \cite{word2vec1,mnih}. In addition, the correlation between model word similarity scores and human similarity judgments is typically high \cite{levy2015improving}. However, no previous work (to our knowledge) evaluates the similarity judgments of character-level models against human annotators.

\section{The Char2Vec model}

\begin{figure}[t]
    \centering
    \includegraphics[width=\columnwidth]{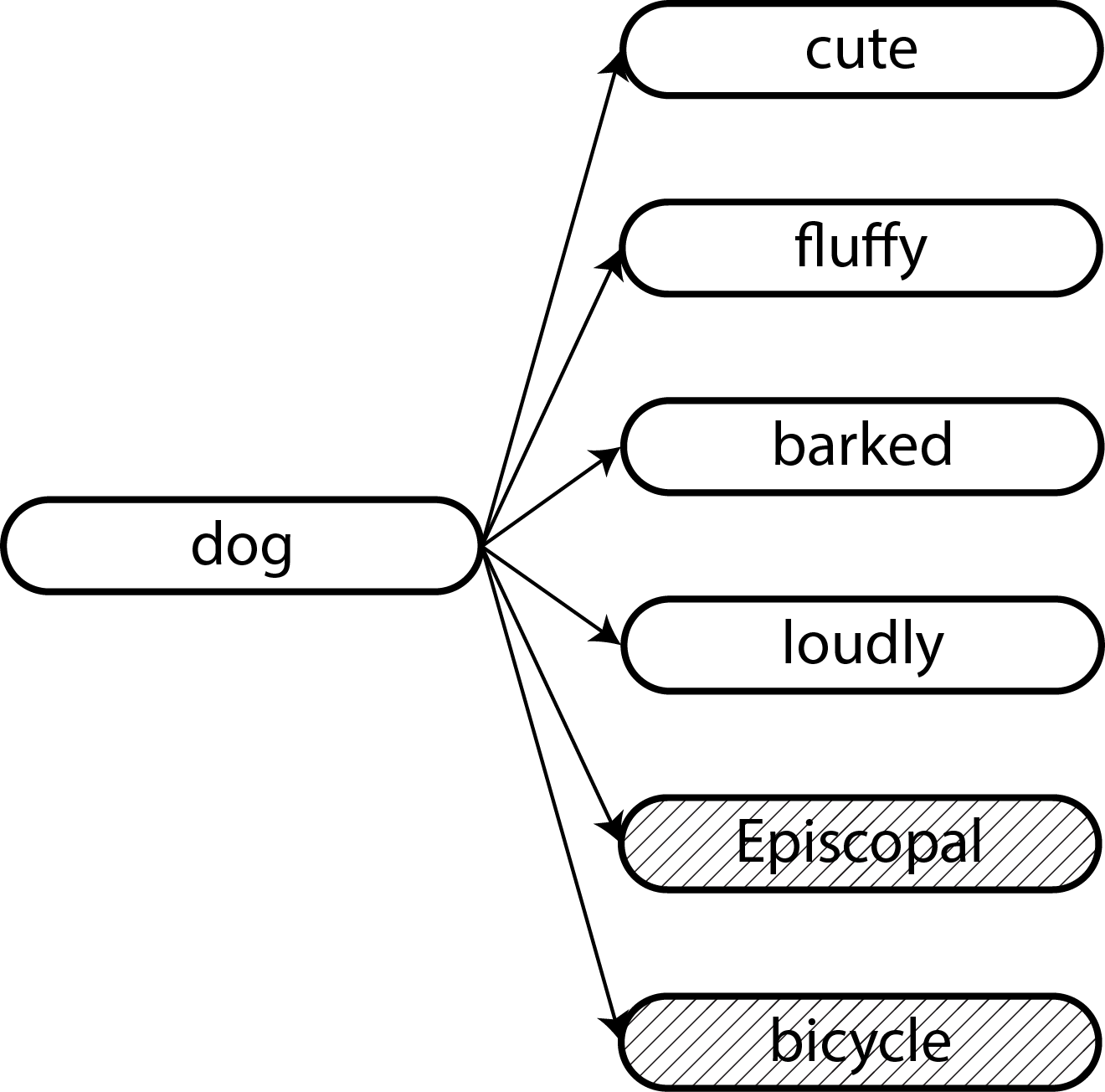}
    \caption{A graphical illustration of SGNS. The target vector for `dog' is learned to have high inner product with the context vectors for words seen in the context of `dog' (no shading), while having low inner product with random negatively sampled words (shaded)}
    \label{fig:SGNS}
\end{figure}

We hypothesize that by incorporating morphological knowledge directly into a character-level model, one can improve the ability of character-level models to learn compositional word semantics. In addition, we hypothesize that incorporating morphological knowledge helps structure the embedding space in such a way that affixation corresponds to a regular shift in the embedding space. We test both hypotheses directly in \S\ref{wordsim} and \S\ref{syntactic} respectively.

The starting point for our model is the skip-gram with negative sampling (SGNS) objective of Mikolov et al. \shortcite{word2vec2}. For a vocabulary $V$ of size $|V|$ and embedding size $N$, SGNS learns two embedding tables $W, C \in \mathbb{R}^{N \times |V|}$, the target and context vectors. Every time a word $w$ is seen in the corpus with a context word $c$, the tables are updated to maximize
\begin{equation}
\label{SGNS}
    \log \sigma(w \cdot c) + \sum_{i = 1}^{k} \mathbb{E}_{\tilde{c}_i \sim P(w)} [\log \sigma(-w \cdot \tilde{c}_i)]
\end{equation}
where $P(w)$ is a noise distribution from which we draw $k$ negative samples. In the end, the target vector for a word $w$ should have high inner product with context vectors for words with which it is typically seen, and low inner products with context vectors for words it is not typically seen with. Figure \ref{fig:SGNS} illustrates this for a particular example. In Mikolov et al. \shortcite{word2vec2}, the noise distribution $P(w)$ is proportional to the unigram probability of a word raised to the 3/4th power \cite{word2vec2}.

Our innovation is to replace $W$ with a trainable function $f$ that accepts a sequence of characters and returns a vector of length $N$ (i.e. $f: A^{<\omega} \to \mathbb{R}^N$, where $A$ is the alphabet we are considering and $A^{<\omega}$ denotes the finite length strings over the alphabet $A$). We still keep the table of context embeddings $C$, and our model objective is still to minimize
\begin{equation}
    \log \sigma(f(w) \cdot c) + \sum_{i = 1}^{k} \mathbb{E}_{\tilde{c}_i \sim P(w)} [\log \sigma(-f(w) \cdot \tilde{c}_i)]
\end{equation}
where we now treat $w$ as a sequence of characters. After training, $f$ can be used to produce an embedding for any sequence of characters, even if it was not previously seen in training.

\begin{figure}[t]
    \centering
    \includegraphics[width=\columnwidth]{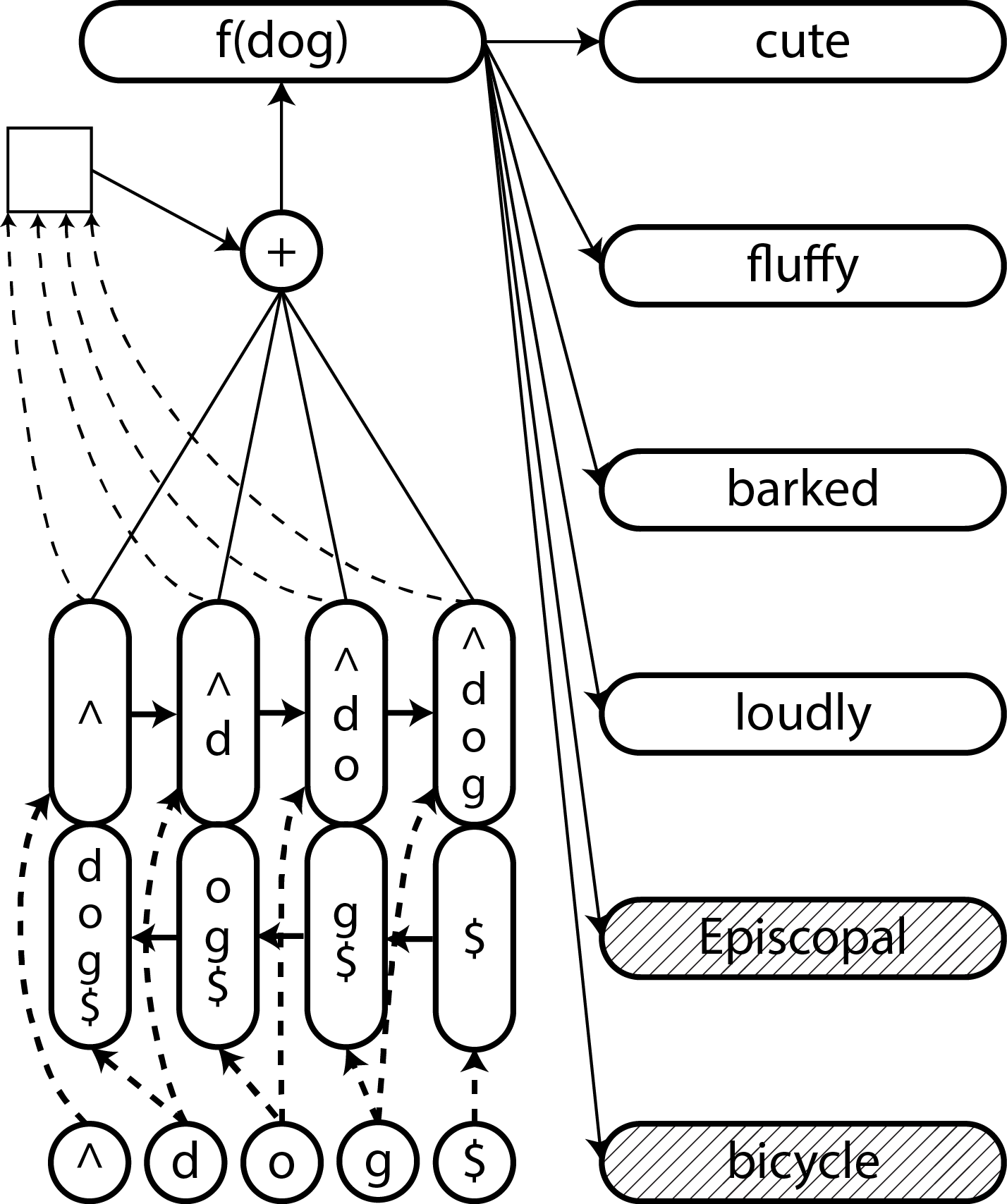}
    \caption{An illustration of Char2Vec. A bidirectional LSTM reads the word (start and end of word symbols represented by \^{} and \$ respectively), outputting a sequence of hidden states. These are then passed through a feed-forward layer (not shown), weighted by an attention model (the square box in the diagram) and summed to obtain the final word representation.}
    \label{fig:char2vec}
\end{figure}

The process of calculating $f$ on a word is illustrated in Figure \ref{fig:char2vec}. We first pad the word with beginning and end of word tokens, and then pass the characters of the word into a character lookup table. As the link between characters and morphemes is non-compositional and requires essentially memorizing a sequence of characters, we use LSTMs \cite{lstm} to encode the letters in the word, as they have been shown to capture non-local and non-linear dependencies. We run a forward and a backward LSTM over the character embeddings. The forward LSTM reads the beginning of word symbol, but not the end of word symbol, and the backward LSTM reads the end of word symbol but not the beginning of word symbol. This is necessary to align the resulting embeddings, so that the LSTM hidden states taken together correspond to a partition of the word into two without overlap.

The LSTMs output two sequences of vectors $h_0^{f}, \dots, h_n^f$ and $h_n^{b}, \dots, h_0^b$. We then concatenate the resulting vectors, and pass them through a shared feed-forward layer to obtain a final sequence of vectors $h_i$. Each vector corresponds to two half-words: one half read by the forward LSTM, and the other by the backward LSTM.

We then learn an attention model over these hidden states: given a hidden state $h_i$, we calculate a weight $\alpha_i = a(h_i)$ such that $\sum \alpha_i = 1$, and then calculate the resulting vector for the word $w$ as $f(w) = \sum \alpha_i h_i$. Following Bahdanau et al. \shortcite{bahdanau}, we calculate $a$ as 
\begin{equation}
a(h_i) = \frac{\exp(v^{T} \tanh(Wh_i))}{\sum_j \exp(v^{T} \tanh(Wh_j))}
\end{equation}
i.e. a softmax over the hidden states.

\subsection{Capturing morphology via attention}
\label{sec:morphology}
\begin{figure}[t]
    \centering
    \includegraphics[width=\columnwidth]{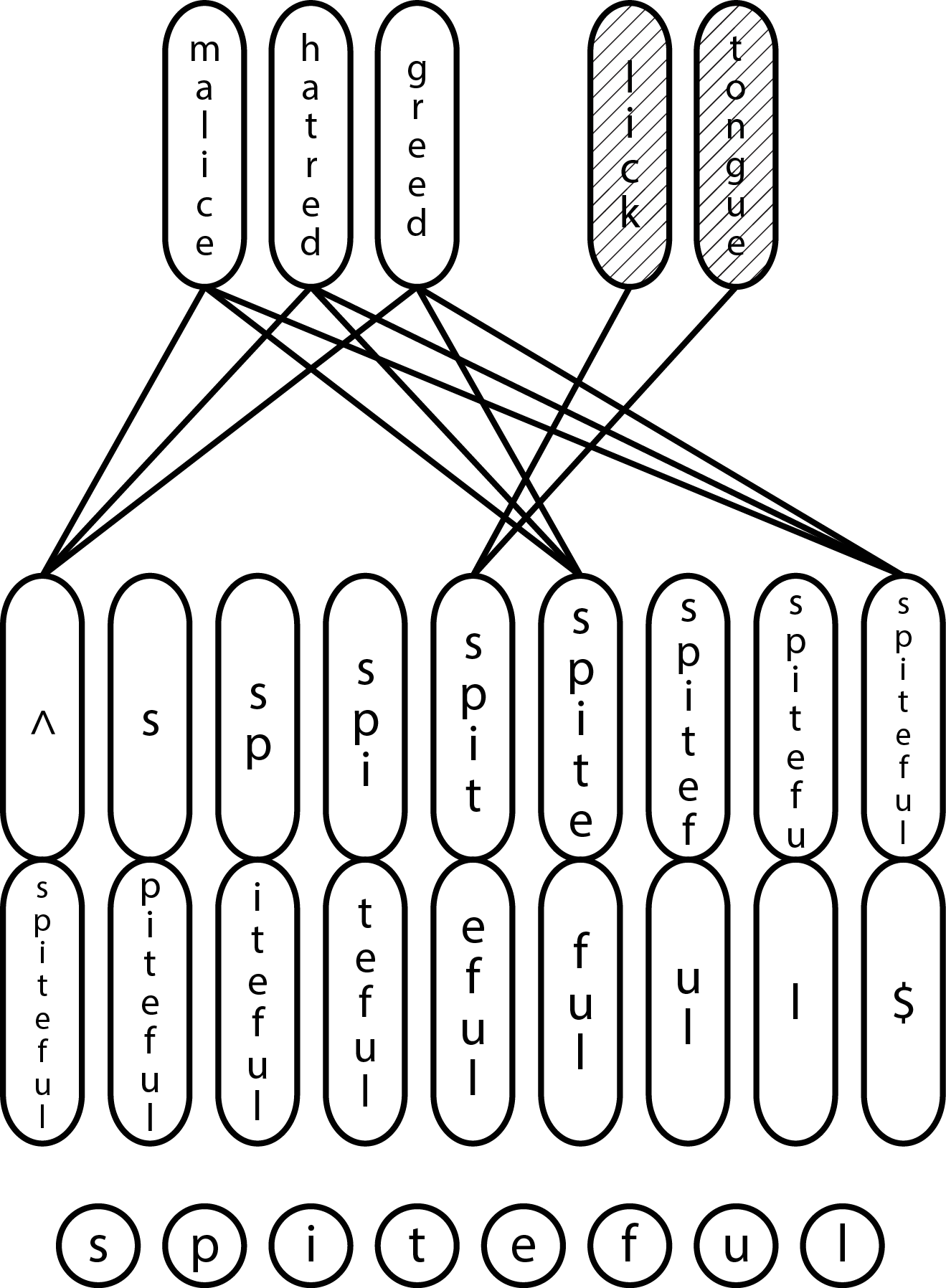}
    \caption{An illustration of the attention model (start and end of word symbols omitted). The root morpheme contributes the most to predicting the context, and is upweighted. In contrast, another potential split is inaccurate, and predicts the wrong context words. This is downweighted.}
    \label{fig:segmentation}
\end{figure}

Previous work on bidirectional LSTM character-level models used both LSTMs to read the entire word \cite{wangling,ballesteros}. This can lead to redundancy, as both LSTMs are used to capture the full word. In contrast, our model is capable of splitting the words and optimizing the two LSTMs for modelling different halves. This means one of the LSTMs can specialize on word prefixes and roots, while the other memorizes possible suffixes. In addition, when dealing with an unknown word, it can be split into known and unknown components. The model can then use the semantic knowledge it has learnt for a known component to predict a representation for the unknown word as a whole.

We hypothesize that the natural place to split words is on morpheme boundaries, as morphemes are the smallest unit of language which carry semantic meaning. We test the splitting capabilities of our model in \S\ref{morphology}.

\section{Experiments}

We evaluate our model on three tasks: morphological analysis (\S\ref{morphology}), semantic similarity (\S\ref{wordsim}), and analogy retrieval (\S\ref{syntactic}). We trained all of the models once, and then use the same trained model for all three tasks -- we do not perform hyperparameter tuning to optimize performance on each task.

We trained our Char2Vec model on the Text8 corpus, consisting of the first 100MB of a 2006 cleaned-up dump of Wikipedia\footnote{available at \texttt{mattmahoney.net/dc/text8}}. We only trained on words which appeared more than 5 times in our corpus. We used a context window size of 3 words either side of the target word, and took 11 negative samples per positive sample, using the same smoothed unigram distribution as \texttt{word2vec}. The model was trained for 3 epochs using the Adam optimizer \cite{adam}. All experiments were carried out using Keras \cite{keras} and Theano \cite{theano1,theano2}. We initialized the context lookup table using \texttt{word2vec}\footnote{We use the Gensim implementation: \texttt{https://radimrehurek.com/gensim/}}, and kept it fixed during training. \footnote{We experimented with updating the initialized context lookup tables, and with randomly initialized context lookups, but found they were influenced too much by orthographic similarity from the character encoder.} In all character-level models, the character embeddings have dimension $d_C = 64$, while the forward and backward LSTMs have dimension $d_{LSTM} = 256$. The concatenation of both therefore has dimensionality $d = 512$. The concatenated LSTM hidden states are then compressed down to $d_{word} = 256$ by a feed-forward layer.

As baselines, we trained a SGNS model on the same dataset with the same parameters. To test how much the attention model helps the character-level model to generalize, we also trained the Char2Vec model without the attention layer, but with the same parameters. In this model, the word embeddings are just the concatenation of the final forward and backward states, passed through a feedforward layer. We refer to this model as \textsc{C2V-NO-ATT}. We also constructed count-based vectors using SVD on PPMI-weighted co-occurence counts, with a window size of 3. We kept the top 256 principal components in the SVD decomposition, to obtain embeddings with the same size as our other models.

\subsection{Morphological awareness}
\label{morphology}

The main innovation of our Char2Vec model compared to existing recurrent character-level models is the capability to split words and model each half independently. Here we test whether our model segmentations correspond to gold-standard morphological analyses.

We obtained morphological analyses for all the words in our training vocabulary which were in the English Lexicon Project \cite{englex}. We then converted these into surface-level segmentations using heuristic affix-matching, and used this as a gold-standard morphemic analysis. We ended up with 14682 words, of which 7867 have at least two morphemes and 1138 have at least three.

Evaluating morphological segmentation is a long-debated issue \cite{orthography}. Traditional hard morphological analyzers are normally evaluated on border $F_1$ -- that is, how many morpheme borders are recovered. However, our model does not actually posit any hard morpheme borders. Instead, it just associates each character boundary with a weight. Therefore, we treat the problem of recovering intra-word morpheme boundaries as a ranking problem. We rank each inter-character boundary of a word according to our model weights, and then evaluate whether our model ranks morpheme boundaries above non-morpheme boundaries.

We use mean average precision (MAP) as our evaluation metric. We first calculate precision at $N$ for each word, until all the gold standard morpheme boundaries have been recovered. Then, we average over $N$ to obtain the average precision (AP) for that word. We then calculate the mean of the APs across all words to obtain the MAP for the model.

We report results of a random baseline as a point of comparison, which randomly places morpheme boundaries inside the word. We also report the results of the Porter stemmer\footnote{We used the NLTK implementation}, where we place a morpheme boundary at the end of the stem, then randomly thereafter.

Finally, we trained Morfessor 2.0\footnote{We used the Python implementation} \cite{morfessor} on our corpus, using an initial random split value of 0.9, and stopping training when the difference in loss between successive epochs is less than 0.1\% of the total loss. We then used our trained Morfessor model to predict morpheme boundaries\footnote{We found Morfessor to be quite conservative by default in its segmentations. The 2nd ranked segmentation gave better MAPs, which are the results we describe.}, and randomly permuted the morpheme boundaries and ranked them ahead of randomly permuted non-morpheme boundaries to calculate MAP.

As the test set is dominated by words with simple morphology, we also extracted all the morphologically rich words with 3 or more morphemes, and created a separate evaluation on this subsection. We report the results in Table \ref{stemming}.

\begin{table}[t]
\centering{
\resizebox{\columnwidth}{!}{
\begin{tabular}{c|cc}
\toprule
     Model & All word MAP & Rich-morphology MAP \\
\midrule
     Random & 0.233 & 0.261 \\
     Porter Stemmer & \textbf{0.705} & 0.446 \\
     Morfessor & 0.631 & 0.500 \\
     Char2Vec & 0.593 & \textbf{0.586} \\
\bottomrule
\end{tabular}
}
}
\caption{Results at retrieving intra-word morpheme boundaries.}
\label{stemming}
\end{table}

As the results show, our model performs the best out of all the methods at analysing morphologically rich words with multiple morphemes. On these words, our model even outperforms Morfessor, which is explicitly designed as a morphological analyzer. This shows that our model learns splits which correspond well to human morphological analysis, even though we build no morphological knowledge into our model. However, when evaluating on all words, the Porter stemmer has a great advantage, as it is rule-based and able to give just the stem of words with great precision, which is effectively giving a canonical segmentation for words with just 2 morphemes.

We show some model analyses against the gold standard in Table \ref{splits}.

\begin{table}[t]
\centering
\resizebox{\columnwidth}{!}{
\begin{tabular}{c|cc}
\toprule
     Word & Model analysis & Gold-standard analysis \\
\midrule
     \textit{carrying} & carry \textbar ing & carry \textbar ing \\
     \textit{leninism} & lenin \textbar ism & lenin \textbar ism \\
     \textit{lesbianism} & lesbia \textbar nism & lesbian \textbar ism \\
     \textit{buses} & buse \textbar s & bus \textbar es \\
     \textit{government} & gove \textbar rnment & govern \textbar ment \\
\bottomrule
\end{tabular}
}
\caption{Some model morphological analyses for words with two morphemes, where we take the highest inter-word weight as the split point for the word.}
\label{splits}
\end{table}

\subsection{Capturing semantic similarity}
\label{wordsim}

\begin{table}[t]
\centering
\resizebox{\columnwidth}{!}{
\begin{tabular}{l|ccc}
\toprule
     Model & WordSim353 & MEN Test & RW \\
\midrule
     \textsc{PPMI-SVD} & 0.607 & \textbf{0.601} & 0.293 \\
     \textsc{SGNS} & \textbf{0.667} & 0.557 & \textbf{0.388} \\
     \textsc{C2V-NO-ATT} & 0.361 & 0.298 & 0.317 \\
     \textsc{CHAR2VEC} & 0.345 & 0.322 & 0.282 \\
\bottomrule
\end{tabular}
}
\caption{Similarity correlations of in-vocabulary word pairs between the models and human annotators.}
\label{in-vocab_sim}
\end{table}

\begin{table}[t]
\centering
\resizebox{\columnwidth}{!}{
\begin{tabular}{l|cccc}
\toprule
     Model & WordSim353 & MEN Test & RW & RW OOV\\
\midrule
     \textsc{C2V-NO-ATT} & \textbf{0.358} & 0.292 & \textbf{0.273} & 0.233 \\
     \textsc{CHAR2VEC} & 0.340 & \textbf{0.318} & 0.264 & \textbf{0.243} \\
\bottomrule
\end{tabular}
}
\caption{Similarity correlations of all word pairs between the character-level models and human annotators. RW OOV indicates results specifically on pairs in the RW dataset with at least one word not seen in the training corpus.}
\label{all-vocab_sim}
\end{table}

\begin{table*}[t]
\centering{
\begin{tabular}{cccccc}
\toprule
     & \multicolumn{3}{c}{In-vocabulary} & \multicolumn{2}{c}{Out-of-Vocabulary} \\
     & \textit{germany} & \textit{football} & \textit{bible} & \textit{foulness} & \textit{definately} \\
     \cmidrule(lr){2-4}
     \cmidrule(lr){5-6}
     \multirow{5}{*}{Char2Vec unfiltered} 
      & \textit{germaine} & \textit{footballer} & \textit{bibles} & \textit{illness} & \textit{definitely} \\
      & \textit{germanies} & \textit{footballing} & \textit{testament} & \textit{seriousness} & \textit{indefinitely} \\
      & \textit{germain} & \textit{footballing} & \textit{librarianship} & \textit{sickness} & \textit{enthusiastically} \\
      & \textit{germano} & \textit{foosball} & \textit{literature} & \textit{loudness} & \textit{emphatically} \\
      & \textit{germaniae} & \textit{footballers} & \textit{librarian} & \textit{cuteness} & \textit{consistently} \\
     \\
     \multirow{5}{*}{Char2Vec filtered}
      & \textit{poland} & \textit{footballer} & \textit{testament} & \textit{illness} & \textit{definitely} \\
      & \textit{german} & \textit{basketball} & \textit{literature} & \textit{blindness} & \textit{consistently} \\
      & \textit{spain} & \textit{tennis} & \textit{hebrew} & \textit{consciousness} & \textit{drastically} \\
      & \textit{germans} & \textit{rugby} & \textit{judaism} & \textit{hardness} & \textit{theoretically} \\
      & \textit{france} & \textit{baseball} & \textit{biblical} & \textit{weakness} & \textit{infinitely} \\
\bottomrule
\end{tabular}
}
\caption{Filtered and unfiltered model nearest neighbours for some in-vocabulary and out-of-vocabulary words}
\label{nearestneighbours}
\end{table*}

Next, we tested our model similarity scores against human similarity judgments. For these datasets, human annotators are asked to judge how similar two words are on a fixed scale. Model word vectors are evaluated based on ranking the word pairs according to their cosine similarity, and then measuring the correlation (using Spearman's $\rho$) between model judgments and human judgments \cite{levy2015improving}.

We use the WordSim353 dataset \cite{wordsim}, the test split of the MEN dataset \cite{MEN}, and the Rare Word (RW) dataset \cite{luong}. The word pairs in the WordSim353 and MEN datasets are typically simple, commonly occurring words denoting basic concepts, whereas the RW dataset contains many morphologically derived words which have low corpus frequencies. This is reflected by how many of the test pairs in each dataset contain out of vocabulary (OOV) items: 3/353 and 6/1000 of the word pairs in WordSim353 and MEN, compared with 1083/2034 for the RW dataset.

We report results for in-corpus word pairs in Table \ref{in-vocab_sim}, and for all word pairs for those models able to predict vectors for unseen words in Table \ref{all-vocab_sim}. 

Overall, word-based embedding models learn vectors that correlate better with human judgments, particularly for morphologically simple words. However, character-based models are competitive with word-based models on the RW dataset. While the words in this dataset appear rarely in our corpus (of the in-corpus words, over half appear fewer than 100 times), each morpheme may be common, and the character-level models can use this information.  We note that on the entire RW dataset (of which over half contain an OOV word), the character-based models still perform reasonably. We also note that on word pairs in the RW test containing at least one OOV word, the full Char2Vec model outperforms the C2V model without morphology. This suggests that character-based embedding models are learning to morphologically analyse complex word forms, even on unseen words, and that giving the model the capability to learn word segments independently helps this process. 

We also present some word nearest neighbours for our Char2Vec model in Table \ref{nearestneighbours}, both on the whole vocabulary and then filtering the nearest neighbours to only include words which appear 100 times or more in our corpus. This corresponds to keeping the top 10k words, which is common among language models \cite{wangling,yoonkim}. We note that nearest neighbour predictions include words that are orthographically distant but semantically similar, showing that our model has the capability to learn to compose characters into word meanings.

We also note that word nearest neighbours seem to be more semantically coherent when rarely-observed words are filtered out of the vocabulary, and more based on orthographic overlap when the entire vocabulary is included. This suggests that for rarely-observed words, the model is basing its predictions on orthographic analysis, whereas for more commonly observed words it can `memorize' the mapping between the orthography and word semantics.

\subsection{Capturing syntactic and semantic regularity}
\label{syntactic}

\begin{table}[t]
\centering
\resizebox{\columnwidth}{!}{
\begin{tabular}{lccc}
\toprule
     Model & All Acc & Sem. Acc & Syn. Acc \\
\midrule
     \textsc{PPMI-SVD} & 0.365 & \textbf{0.444} & 0.341 \\
     \textsc{SGNS} & \textbf{0.436} & 0.339 & 0.513 \\
     \textsc{C2V-NO-ATT} & 0.316 & 0.016 & 0.472 \\
     \textsc{CHAR2VEC} & 0.355 & 0.025 & \textbf{0.525} \\
\bottomrule
\end{tabular}
}
\caption{Results on the Google analogy task}
\label{googleanalogy}
\end{table}

Finally, we evaluate the structure of the embedding space of our various models. In particular, we test whether affixation corresponds to regular linear shifts in the embedding space.

To do this, we use the Google analogy dataset \cite{word2vec1}. This consists of 19544 questions of the form ``\textit{A} is to \textit{B} as \textit{C} is to \textit{X}''. We split this collection into semantic and syntactic sections, based on whether the analogies between the words are driven by morphological changes or deeper semantic shifts. Example semantic questions are on capital-country relationships (``\textit{Paris} is to \textit{France} as \textit{Berlin} is to \textit{X}) and currency-country relationships. Example syntactic questions are adjective-adverb relationships (``\textit{amazing} is to \textit{amazingly} as \textit{apparent} is to \textit{X}'') and opposites formed by prefixing a negation particle (``\textit{acceptable} is to \textit{unacceptable} as \textit{aware} is to \textit{X}''). This results in 5537 semantic analogies and 10411 syntactic analogies.

We use the method of Mikolov et al. \shortcite{word2vec1} to answer these questions. We first $\ell_2$-normalize all of our word vectors. Then, to answer a question of the form ``\textit{A} is to \textit{B} as \textit{C} is to \textit{X}'', we find the word $w$ which satisfies
\begin{equation}
    w = \argmax_{w \in V - \{a, b, c\}} \cos(w, b - a + c)
\end{equation}
where $a,\, b,\, c$ are the word vectors for the words \textit{A}, \textit{B} and \textit{C} respectively.

We report the results in Table \ref{googleanalogy}. The most intriguing result is that character-level models are competitive with word-level models for syntactic analogy, with our Char2Vec model holding the best result for syntactic analogy answering. This suggests that incorporating morphological knowledge explicitly rather than latently helps the model learn morphological features. However, on the semantic analogies, the character-based models do much worse than the word-based models. This is perhaps unsurprising in light of the previous section, where we demonstrate that character-based models do worse at the semantic similarity task than word-level models.

\section{Discussion}

We only report results for English. However, English is a morphologically impoverished language, with little inflection and relatively few productive patterns of derivation. Our morphology test set reflects this, with over half the words consisting of a simple morpheme, and over 90\% having at most 2 morphemes.

This is unfortunate for our model, as it performs better on words with richer morphology. It gives consistently more accurate morphological analyses for these words compared to standard baselines, and matches word-level models for semantic similarity on rare words with rich morphology. In addition, it seems to learn morphosyntactic features to help solve the syntactic analogy task. Most of all, it is language-agnostic, and easy to port across different languages. We thus expect our model to perform even better for languages with a richer morphology than English, such as Turkish and German.

\section{Conclusion}

In this paper, we present a model which learns morphology and word embeddings jointly. Given a word, it splits the word in to segments and ranks the segments based on their context-predictive power. Our model can segment words into morphemes, and also embed the word into a representation space.

We show that our model is competitive at the task of morpheme boundary recovery compared to a dedicated morphological analyzer, beating dedicated analyzers on words with a rich morphology. We also show that in the representation space word affixation corresponds to linear shifts, demonstrating that our model can learn morphological features.

Finally, we show that character-level models, while outperformed by word-level models generally at the task of semantic similarity, are competitive at representing rare morphologically rich words. In addition, the character-level models can predict good quality representations for unseen words, with the morphologically aware character-level model doing slightly better.

\bibliography{acl2016}
\bibliographystyle{acl2016}
\end{document}